\documentclass[letterpaper]{article} 
\usepackage{aaai25}  
\usepackage{times}  
\usepackage{helvet}  
\usepackage{courier}  
\usepackage[hyphens]{url}  
\usepackage{amsmath} 
\usepackage{graphicx} 
\usepackage{natbib}  
\usepackage{caption} 
\usepackage{makecell}
\usepackage{float}

\usepackage{algorithm}
\usepackage{algorithmic}
\usepackage{newfloat}
\usepackage{listings}
\DeclareCaptionStyle{ruled}{labelfont=normalfont,labelsep=colon,strut=off} 
\floatstyle{ruled}
\newfloat{listing}{tb}{lst}{}
\floatname{listing}{Listing}
\nocopyright
\title{MAR: Multi-Agent Reflexion Improves Reasoning Abilities in LLMs}
\author{
    Onat Ozer\textsuperscript{\rm 1}, Yuchen Wang\textsuperscript{\rm 1}, Grace Wu\textsuperscript{\rm 1}, Daniel Dosti\textsuperscript{\rm 1}, Honghao Zhang\textsuperscript{\rm 1}, Vivi De La Rue\textsuperscript{\rm 1}\\
}
\affiliations{
    \textsuperscript{\rm 1}University of Michigan\\
    \{ozeronat, wangyuch, gracejwu, ddosti, 
    honghaoz, vdelarue\}@umich.edu\\
}

\begin{document}

\maketitle
\begin{abstract}

Large Language Models (LLMs) have shown strong capabilities in reasoning and program synthesis, yet enabling them to improve through experience remains a central challenge. Reinforcement learning offers one opportunity for agent improvement, but updating large models is computationally expensive and requires significant environment interaction. The Reflexion framework introduced a lightweight alternative in which an LLM agent generates natural language reflections about its failures and stores them as episodic memory that guides future attempts. Reflexion demonstrated meaningful gains on multi hop question answering and code generation without modifying model parameters. However, our replication shows that the single agent design of Reflexion leads to systematic shortcomings. The same model generates actions, evaluates its own behavior, and produces reflections, which often results in repeated reasoning errors, confirmation bias, and limited corrective feedback.

To address these limitations, we reproduce Reflexion on HotPotQA and HumanEval using GPT 3.5 and GPT 4, and we analyze failure modes through detailed logging of trajectories and reflections. We consistently observe that self reflections tend to repeat earlier misconceptions and do not introduce new reasoning paths on difficult examples. Motivated by these findings, we introduce Multi Agent Reflexion (MAR), a structured multi agent extension that incorporates diverse reasoning personas and a judge model that synthesizes their critiques into a unified reflection. MAR separates the processes of acting, diagnosing, critiquing, and aggregating, which reduces shared blind spots and prevents the repeated reinforcement of earlier mistakes. On both benchmarks, MAR improves directly over Reflexion. MAR increases HotPotQA exact-match accuracy by 3 points, from 44 to 47, and improves HumanEval pass@1 from 76.4 to 82.6, a 6.2-point gain. MAR also reduces the stagnation observed in single agent Reflexion, indicating that multi agent interaction offers a simple and effective path toward more reliable LLM agents.


\end{abstract}

\section{Introduction}

Large Language Models (LLMs) has evolved to generate strong reasoning traces and high-quality code, but enabling them to improve through experience remains an open problem. Traditional reinforcement learning offers one mechanism for agent improvement, but updating billion-parameter policies is computationally costly and data-intensive. In-context learning \cite{brown:20_fewshot}, where an LLM adapts using examples directly in the prompt, offers one path toward improving model performance without additional training. Building on this idea, prompting strategies such as chain-of-thought (CoT), expert-style prompting, and ReAct have been shown to enhance LLM reasoning and decision-making.


 Reflexion \cite{shinn2023reflexion} integrates these advances while adding a mechanism for the model to self-reflect on its own mistakes: after each attempt, the agent converts feedback into natural-language guidance for the next trial, analogous to how reinforcement learning algorithms use reward signals to update a policy \cite{sutton_barto_2018}.

Reflexion reports substantial gains over baseline LLM and CoT/ReAct agents on HotPotQA \cite{yang2018hotpotqa} and HumanEval \cite{chen2021codex}, improving exact-match accuracy by roughly 20 points on HotPotQA and pass@1 accuracy by 11 points on HumanEval. We focus on these benchmarks for replication since \textit{multi-hop question} (questions where a model must combine information across several steps of retrieveal) answering and program synthesis are the two high-impact domains where iterative self-improvement most meaningfully enhance LLM agent performance, and covers the majority of use cases. 

Despite the gains of the Reflexion framework, in our replication, we consistently observed that Reflexion is vulnerable to degeneration-of-thought \cite{liang2023divergent}, where the agent repeats the same flawed reasoning across iterations even when explicit failures are identified. To mitigate this limitation, we introduce \textbf{Multi-Agent Reflexion (MAR)}, a framework that replaces single-agent self-critique with a structured debate among diverse persona-based critics. These multiple critics generate richer reflections that more effectively guide the agent toward improved strategies.

Our full codebase, including replication scripts, logging instrumentation, and the Multi-Agent Reflexion implementation, is available at:
\url{https://github.com/danpro1011/reflexion}.

\subsection{Related Work}
Recent research explores improving LLM reasoning through self-generated feedback, iterative refinement, and multi-agent interaction. Self-critique methods such as Self-Refine and Reflexion show that verbal feedback can improve performance without updating model weights, though they often rely on a single agent with fixed reasoning patterns. Multi-agent debate introduces structured disagreement and persona diversity, helping reduce shared blind spots; our approach builds on this idea by integrating diverse critic personas into an episodic memory framework that supports more stable and transferable improvements across tasks.

\noindent\textbf{Self-Generated Feedback and Iterative Refinement:}
Several works explore letting LLMs critique and revise their own outputs. Self-Refine, for example, iteratively generates output, asks the model to provide targeted feedback, and then uses that feedback to produce a revised answer \cite{madaan2023selfrefine}. 

\noindent\textbf{Multi-Agent Debate and Divergent Reasoning:}  
Multi-agent debate methods instantiate multiple LLM agents that propose answers, critique each other's reasoning, and then aggregate the resulting arguments into a final decision \cite{du2023multiagentdebate}. Recent work shows that structured disagreement can improve factual accuracy and robustness on challenging reasoning benchmarks by promoting divergent reasoning and reducing shared blind spots \cite{liang2023divergent}. Other multi-agent frameworks introduce persona-driven or role-based heterogeneity among agents and coordinate their outputs through majority voting or a centralized judge model \cite{li2023camel}. 


\noindent\textbf{Chain-of-Thought and Tool-Augmented Reasoning:}
Chain-of-thought (CoT) prompting shows that LLMs can solve substantially harder reasoning problems when encouraged to generate intermediate natural-language steps before answering \cite{wei2022cot}. CoT is purely a single-pass prompting strategy. The model produces one reasoning trace and does not explicitly learn across episodes. ReAct extends this idea by interleaving reasoning tokens with actions, so that an LLM can plan, call tools or APIs, and react to new observations in text-based environments \cite{yao2022react}.

 
\section{Background}

\subsection{Reflexion Framework}

Reflexion \cite{shinn2023reflexion} wraps a language model in a simple loop that turns past failures into natural-language guidance for future attempts. Conceptually, the framework decomposes the agent into three prompted roles plus an episodic memory.

\noindent\textbf{Actor ($M_a$).}
The Actor is the LLM prompted to behave as the agent’s policy. It generates actions, intermediate reasoning, and final answers. Reflexion is compatible with multiple prompting styles; the original paper focuses on:
\begin{itemize}
    \item \emph{Chain-of-Thought (CoT) \cite{wei2022cot}} prompting, which encourages structured step-by-step reasoning.
    \item \emph{ReAct \cite{yao2022react}}, which interleaves reasoning steps with explicit actions taken in an external environment.
\end{itemize}

\noindent\textbf{Evaluator ($M_e$).}
The Evaluator judges whether an Actor trajectory successfully solves the task. Depending on the domain, $M_e$ may be another LLM prompt, a heuristic rule, or an external execution environment such as code unit tests. Importantly, Reflexion only requires a coarse signal, such as success/failure, not a dense numeric reward.

\noindent\textbf{Self-Reflector ($M_{sr}$).}
When the Evaluator indicates failure, the Self-Reflection model converts this outcome into verbal feedback. Given the failed trajectory and its evaluation, $M_{sr}$ produces natural-language reflections that diagnose what went wrong, explain why the reasoning failed, and suggest strategies to try the next attempt. These ``verbal rewards'' play the role of a lightweight optimization signal.

\begin{figure}
    \centering
    \includegraphics[width=0.75\linewidth]{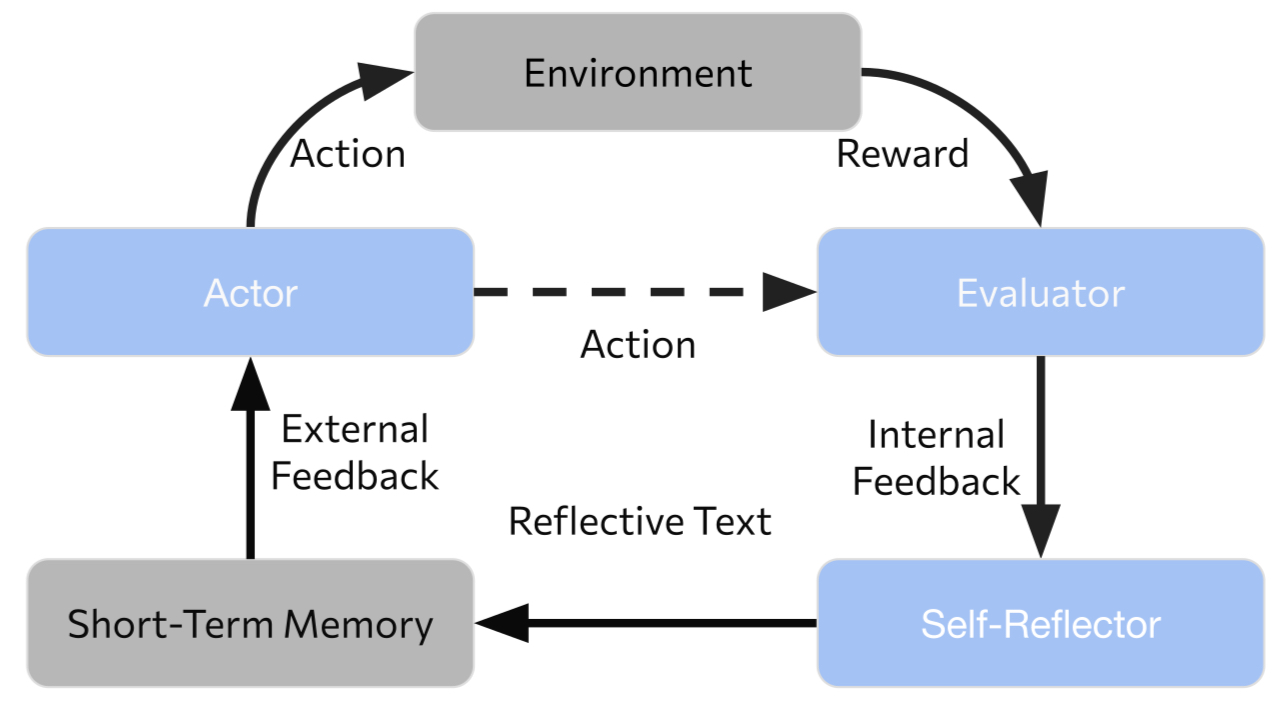}
    \caption{\textbf{The Reflexion Architecture.} Actor performs an action on the environment. The evaluator gives feedback to reflector LLM, with feedback stored in short-term memory. A new action is then performed starting a new iteration.}
    \label{fig:placeholder}
\end{figure}


\noindent At a high level, each failed attempt produces a new reflection that is added to memory, and the Actor’s next attempt is conditioned on this accumulated verbal history. Over multiple trials, the agent can therefore improve its decisions without any parameter updates.

\subsection{Datasets}
We evaluate our approach on two benchmarks that stress different aspects of LLM reasoning. HotPotQA evaluates multi-hop question answering over Wikipedia, requiring models to combine evidence across documents. HumanEval tests program synthesis by checking whether generated functions pass hidden unit tests. Together, these datasets provide complementary settings for analyzing how reflection influences both symbolic reasoning and code generation.



\begin{itemize}

\item \textbf{HotPotQA \citep{yang2018hotpotqa}} is a large-scale question answering benchmark constructed from Wikipedia and designed to test multi-hop reasoning (questions system has to reason with information taken from more than one document to arrive at the answer. Each example provides a question, a set of candidate paragraphs, supporting facts, and a short answer. Successful solutions typically require integrating evidence from more than one source, which makes HotPotQA a natural environment for studying how Reflexion’s episodic memory and reflections influence multi-step reasoning. Following the evaluation protocol in \citet{shinn2023reflexion}, we report Exact Match (EM) on the development set after applying standard normalization procedures such as lowercasing and punctuation removal.


\item \textbf{HumanEval \citep{chen2021codex}} is a benchmark for program synthesis in which each task specifies a function signature, a natural-language description of the intended behavior, and a series of hidden unit tests. A model succeeds if its generated function passes all tests when executed in a sandboxed environment. HumanEval is the domain where the original Reflexion paper observes its largest performance gains relative to GPT-4 baselines, and it provides a clear setting for examining how reflection-guided retries alter the structure and quality of code-generation trajectories.
\end{itemize}

\subsection{Performance Metrics}
We evaluate both the original Reflexion framework and our multi-agent extension using task-appropriate accuracy metrics. For HotPotQA, we report Exact Match (EM), and for HumanEval, we use pass@1 based on whether the generated solution passes all hidden unit tests.

    
    

\begin{itemize}
\item \textbf{Exact Match (EM) for HotPotQA.} EM measures whether a predicted answer exactly matches the ground truth HotPotQA answer matches after applying HotPotQA’s standard normalization steps, including lowercasing and removal of punctuation and articles. 

\item \textbf{pass@1 for HumanEval.} A model receives credit for a HumanEval problem if its single sampled solution passes all hidden unit tests when executed in the official sandboxed environment. 

\end{itemize}



\section{Methodology, Results, and Discussion}

\subsection{Methodology}
We evaluate Reflexion and our extensions on two benchmarks: multi-hop question answering with HotPotQA and program synthesis with HumanEval. Across both settings, we follow the original Reflexion protocol while adding detailed logging to analyze how reasoning, evaluation, and reflection evolve across attempts.

\paragraph{HotPotQA Setup.}
We evaluate Reflexion on the multi-hop HotPotQA benchmark by reproducing the three agent configurations examined in the original study: Baseline GPT-3.5-Turbo using the ReAct framework, and GPT-3.5-turbo using the ReAct-Reflect framework. These frameworks are run on a curated selection of 100 difficult HotPotQa questions, matching the dataset used in their paper. We run these experiments over the course of 5 trials where after each trial, the ReAct agent will try again after reflecting over previous failures. 

\paragraph{HumanEval Setup.}
For the HumanEval benchmark, we replicate the Reflexion pipeline using the official sandboxed execution environment. Each generated solution is executed against the hidden unit tests, and performance is reported using the standard pass@1 metric. Following the original Reflexion protocol, each problem receives one initial attempt followed by a single reflection-guided retry.

\subsection{Results}
This subsection presents our replication results for both datasets, comparing them against the original Reflexion paper. Because our goal is to faithfully reproduce the core experiments, we report the same metrics: Exact Match (EM) for HotPotQA and pass@1 for HumanEval. We include the original scores for reference and leave placeholders for our HotPotQA numbers, which will be filled in after full evaluation.

\subsubsection{HotPotQA Replication}
\label{sec:hotpot-results}

We evaluate Reflexion on HotPotQA using the same baselines as the core paper
(ReAct and Reflexion agents) over the course of 5 trials. Exact Match (EM) scores for the
original paper and our replication are summarized in Table~\ref{tab:hotpot}.

\begin{table}[!ht]
\centering
\scriptsize
\setlength{\tabcolsep}{4pt}
\begin{tabular}{lccc}
\hline
\textbf{HotPotQA (EM)} &
\textbf{ReAct} &
\textbf{Reflexion} \\
\hline
Reflexion Orig. & 34 & 51  \\
Our Replication & \textit{32} & \textit{44} &  \\
\hline
\end{tabular}
\caption{Exact Match (EM) on HotPotQA: original Reflexion numbers vs.\ our replication at the 5th trial.}
\label{tab:hotpot}
\end{table}

\subsubsection{HumanEval Replication}
\label{sec:humaneval-results}
Our experiments used GPT-3.5 Turbo as the previous SOTA baseline and GPT-4 as the current SOTA baseline, matching the setup in the original paper. For each HumanEval function, we allowed a maximum of two trials: an initial attempt, followed by one reflection and one retry, consistent with the original Reflexion protocol for limited-budget code generation.

\begin{table}[h]
\centering
\scriptsize
\setlength{\tabcolsep}{2pt}
\begin{tabular}{lccc}
\hline
\textbf{HumanEval} &
\makecell{\textbf{Prev SOTA}\\\textbf{Pass@1 (GPT-3.5)}} &
\makecell{\textbf{SOTA}\\\textbf{Pass@1 (GPT-4)}} &
\makecell{\textbf{Reflexion}\\\textbf{Pass@1 (GPT-4)}} \\
\hline
Reflexion Orig. & 65.8 & 80.1 & 91.0 \\
Our Replication & 67.1 & 81.7 & 89.4 \\
\hline
\end{tabular}
\caption{Pass@1 comparison between the original Reflexion programming results and our replication.}
\label{tab:replication_comparison}
\end{table}

\section{Extensions}
In this section, we introduce \textbf{Multi-Agent Reflexion (MAR)}, our extension to the Reflexion framework that replaces single-agent self-critique with a coordinated team of persona driven critics meant to address the core failure modes we observed in replication.

\subsection{Motivation}
LLMs frequently struggle to revise their own reasoning, and even with external feedback they tend to reinforce the same flawed chain of thought across successive attempts. This phenomenon is described as degeneration-of-thought \citep{liang2023divergent} or the mental set problem \citep{liu:25}. A central cause of this issue is that self-reflective approaches depend on fixed internal thinking patterns. Even when agents are prompted with different personas, their underlying reasoning strategies often remain uniform. This uniformity prevents the model from exploring alternative perspectives and makes it difficult to escape entrenched errors. These limitations are reflected in the behavior of methods such as Reflexion \citep{shinn2023reflexion} and related approaches \citep{kim:23, madaan2023selfrefine}, which frequently plateau on complex reasoning tasks and can sometimes reinforce earlier mistakes instead of correcting them \citep{huang:24, stechly:23, valmeekam:23}.

In our replication of Reflexion, these limitations manifested consistently. The single-agent loop exhibited two recurring failure modes:

\begin{enumerate}

\item \textbf{Confirmation Bias in Self-Reflection.}
The Actor frequently committed an initial reasoning error. The Evaluator, drawn from the same model family, then attempted to critique the trajectory. However, instead of isolating the root cause, the reflection often:
\begin{itemize}
\item Restated the incorrect reasoning.
\item Justified the error through additional flawed logic.
\item Produced superficial edits.
\end{itemize}

\item \textbf{Mode Collapse Across Iterations.}
Because the same model generated, evaluated, and revised the reasoning, the corrections tended to follow a narrow and repetitive template. In HumanEval, we observed multiple instances in which the system reproduced:
\begin{itemize}
\item Identical buggy loop structures.
\item Repeated indexing mistakes.
\item The same off-by-one error.
\end{itemize}
\end{enumerate}

With a single model acting as its own generator, evaluator, and critic tends to reproduce the same reasoning structure across iterations with little correction. As such, our proposed Multi-Agent Reflexion (MAR) framework explicitly address these weakness by incorporating deliberately varied reasoning strategies into the reflection loop, with a judge that takes away as much meaningful conclusion as possible from each debate. 


\subsection{Extension: Multi-Agent Debate as Reflection}
\begin{figure}[H]
    \centering
    \includegraphics[width=0.75\linewidth]{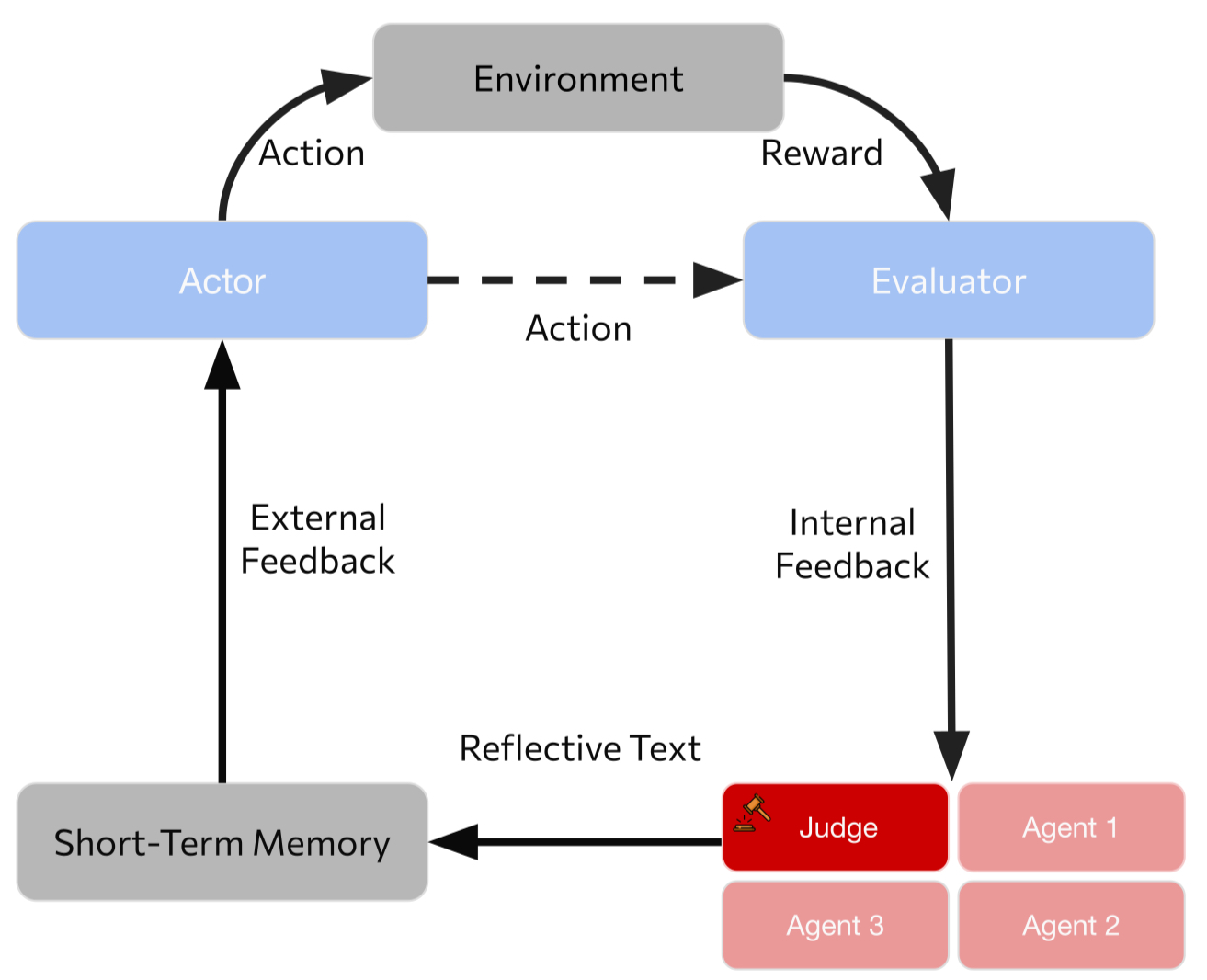}
    \caption{\textbf{The Multi-Agent Reflexion (MAR) Architecture.} This high-level diagram illustrates the extension of the single-agent Reflexion framework.}
    \label{fig:placeholder}
\end{figure}

Our extension replaces the single self-reflecting model in Reflexion with a group of LLM agents that each serve as distinct critics. When the Actor produces an incorrect answer, the system does not rely on a single reflection. Instead, it initiates a structured multi-agent debate in which several persona-guided critics analyze the failed reasoning from different perspectives. Each critic contributes alternative hypotheses, highlights potential flaws, and proposes corrective strategies. A debate coordinator then aggregates these perspectives into a final consensus reflection, which is appended to the Actor’s memory for the next attempt.

The Multi-Agent Reflexion loop works as follows: 

\begin{enumerate}
  \item \textbf{Actor Attempt:} the Actor first attempts the question using the same baseline model as in standard Reflexion.
  \item \textbf{Evaluation:} If incorrect, the system triggers the debate module. Note that this uses the same evaluator as the original Reflexion framework.
  \item \textbf{Initial Diagnosis:} The failed thoughts (scratchpad) of the actor are passed to the judge. The judge asks each persona to write a diagnosis as to why the actor failed the task. 
  \item \textbf{Debates:} After each persona has made a diagnosis, the judge asks whether other personas agree or disagree and refine their critiques.
  \item \textbf{Consensus and Reflection} The judge synthesizes the debate into a single, actionable "Consensus Reflection" which includes error outputs, debate logs, and consensus between debaters depending on the task. If there has been no consensus reached between the debaters, then the personas debate for another round (maximum 2 rounds).
  \item \textbf{Retry} This reflection is injected into the Actor’s memory, guiding the next attempt with specific advice on how to avoid previous pitfalls.
\end{enumerate}

\subsection{Personas in Multi-Agent Debate}

A key component of MAR is the use of intentionally diverse critic personas. We designed personas in a systemic way such that reasoning tendencies differ, inspired by the Society of Mind framework and the divergent-thinking objectives of Multi-Agent Debate (MAD) (Liang et al., 2024). 

We define the following axes to vary across for persona creation. For any given dataset, we carefully craft a set of persona to cover a comprehensively cover these these axes. 

\begin{itemize}
    \item \textbf{Evidence Exploitation:} how strongly the critic demands
    explicit textual support for actions.
    \item \textbf{Exploration:} whether the critic tries to actively pushes the debate toward alternative hypotheses.
    \item \textbf{Specification strictness:} how much the critic optimizes for exact compliance with the metric.
\end{itemize}

Within this design space, we instantiate the following critics
(see Appendix~C for full prompts):

\begin{itemize}
    \item \textbf{Verifier} (high exploit, low exploration, medium strictness)  
    chains of thought.
    \item \textbf{Skeptic} (low exploit, high exploration, medium strictness)
    tuned to counter confirmation bias.
    \item \textbf{Logician} (high exploit, low exploration, high strictness)  
    \item \textbf{Creative} (low exploit, high exploration, high exploration) 
\end{itemize}

 Concrete definitions of personas are given in the appendix (see appendix item C), along with exact prompt used and reasoning behind personas. As future work, we may ask an LLM to generate personas on these axes for a specific dataset.




\subsection{Experimental Setup}
To evaluate our Multi-Agent Reflexion (MAR), our experimental configuration mirrors the structure of the original Reflexion framework where possible, differing only in replacing the single-agent reflection step with a structured multi-agent debate. We use GPT-3.5 Turbo for all agent roles (actor, debaters, and judge) which allows us to isolate the impact of persona-guided multi-agent reasoning while preserving Reflexion’s core design.




Across both HotPotQA and HumanEval, we designed our agent personas to reflect the dominant failure modes observed during replication. For HotPotQA, we selected personas focused on factual grounding, logical consistency, alternative hypothesis generation, and error checking. For code generation, we used personas mirroring real world software roles (e.g., Senior Engineer, QA Engineer, Code Reviewer) to capture algorithmic and specification considerations. Personas were engineered to disagree naturally and expose different lines of reasoning, so that debates surface diverse critiques rather than stylistic variations.

We cap debates at two rounds, based on empirical logs and prior MAD work showing that nearly all meaningful disagreements arise within the first two exchanges.  Trial budgets were also kept intentionally small (five trials for HotPotQA and three for HumanEval), to keep API costs lower. While increasing the limit did yield marginal improvements, we observed sharply diminishing returns beyond these thresholds, accompanied by rapidly increasing API cost and latency. This setup therefore preserves the core design principles of Reflexion while isolating the specific contribution of multi-agent reasoning.
\subsection{Results}


\begin{center}
\captionsetup{type=table}
\scriptsize
\setlength{\tabcolsep}{2pt}
\begin{tabular}{lccc}
\hline
\textbf{} &
\makecell{\textbf{Baseline}\\\textbf{EM (ReAct, GPT 3.5)}} &
\makecell{\textbf{Reflexion + ReAct}\\\textbf{EM (GPT 3.5)}} &
\makecell{\textbf{MAR}\\\textbf{Pass@1 (GPT-3.5)}} \\
\hline
Our Results & 32.0 & 44.0 & \textbf{47.0 }\\
\hline
\end{tabular}
\captionof{table}{EM accuracy on HotPotQA for GPT-3.5, comparing our reproduced Reflexion results and MAR.}
\label{tab:extension-comparison_hotpot}
\end{center}

\paragraph{HotPotQA Results}
As shown in \ref{tab:extension-comparison_hotpot}, our reproduced ReAct baseline matches the Reflexion paper’s reported 32\% EM. Reflexion + ReAct improves this to 44\% EM using GPT-3.5, and MAR out performs both, with an EM of 47\%. While this performance does demonstrate an improvement, we note that this improvement is less than what we would expect, especially compared to our programming results, and attribute this to the shortcomings of the EM metric used for HotPotQA.  

Similar to any other reinforcement learning based algorithm, the robustness of the reward signal is crucial for proper policy improvement. In our analysis of failure cases, we found that EM frequently marks clearly correct answers as incorrect due to formatting sensitivity (e.g., minor differences in spacing, ordering, punctuation, or perfectly valid synonym substitutions). As shown in Appendix A and B, MAR often produced correct reasoning but was penalized for minor surface-level differences. We also show in Appendix E an example reasoning trace where the model initially guesses a semantically correct response, but receives misleading advice from the environment that this result is incorrect, leading it to eventually output the semantically wrong result in response to its reflections. More forgiving metrics (e.g., F1 or semantic-match scoring) would likely reveal stronger MAR gains.

\begin{figure}[h]
    \centering
    \includegraphics[width=0.8\linewidth]{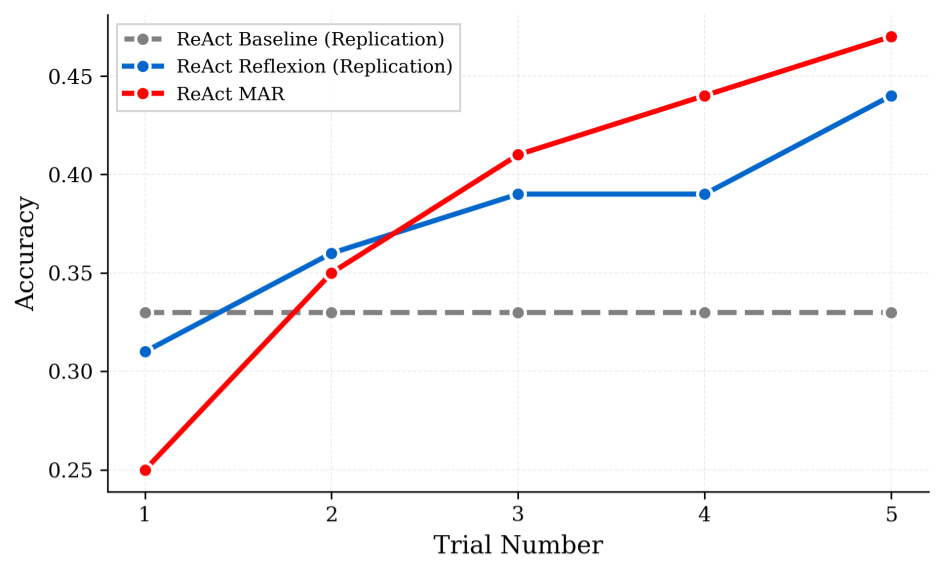}
    \caption{\textbf{Comparison of HotPotQA performance across trials for ReAct, Reflexion, and Multi-Agent Reflexion (MAR)}.
The baseline gpt-3.5-Turbo (grey) and the Reflexion replication results (blue) are shown. MAR (red) offers the highest EM improvement.}
    \label{fig:placeholder}
\end{figure}


\begin{center}
\captionsetup{type=table}
\scriptsize
\setlength{\tabcolsep}{2pt}
\begin{tabular}{lcccc}
\hline
\textbf{} &
\makecell{\textbf{Baseline}\\\textbf{Pass@1 (GPT-3.5)}} &
\makecell{\textbf{Baseline}\\\textbf{Pass@1 (GPT-4)}} &
\makecell{\textbf{Reflexion}\\\textbf{Pass@1 (GPT-3.5)}} &
\makecell{\textbf{MAR}\\\textbf{Pass@1 (GPT-3.5)}} \\
\hline
 & 67.1 & 81.7 & 76.4 & \textbf{82.6}\\
\hline
\end{tabular}
\captionof{table}{Pass@1 results for HumanEval comparing baseline, Reflexion, and MAR.}
\label{tab:extension-comparison_humanEval}
\end{center}

\paragraph{HumanEval Results}
Table \ref{tab:extension-comparison_humanEval} shows that Reflexion improves GPT-3.5 performance from 67.1\% to 76.4\% pass@1, while MAR further increases performance to 82.6\%. 

The pattern reflects how each method handles errors: the baseline often repeats the same mistake, Reflexion corrects it but often gets stuck in a single mode of reasoning, while MAR consistently escapes these failure modes by generating diverse critiques and higher-quality fixes. Overall, the results demonstrate that adding structured multi-agent persona debate yields more reliable corrections and leads to a substantial gain. 

\subsection{Takeaways}

\paragraph{Reduced Degeneration of Thought.}
Single-agent Reflexion often suffers from ``degeneration of thought,'' where the model repeats the same flawed reasoning across retries. Because the Actor, Evaluator, and Reflector are all the same model, its reflections tend to reinforce the original misconception rather than correct it. MAR mitigates this by introducing multiple agents with distinct reasoning styles. Their disagreements surface alternative interpretations of the specification and error trace, helping the system escape the mental-set failures that limit Reflexion.

\paragraph{Higher-quality revisions through aggregation.}
Each MAR trial ends with a judge model that selects the most coherent and spec-aligned proposal from the debate. In practice, agents catch different issues. One may focus on specification accuracy, another on edge cases, another on logic or style. The judge synthesizes these perspectives, producing patches that are more reliable than the single-agent self-reflection loop.

\paragraph{Limitations}
While MAR consistently outperforms single-agent Reflexion, this improvement comes with a significant computational cost. Each failed attempt triggers multiple critic responses, a structured debate, and a judge synthesis step, expanding both token usage and context length. In practice, this leads to roughly a 3x increase in API calls and latency compared to Reflexion. As a result, MAR is substantially more expensive to run, and its scalability is constrained when deployed on large benchmarks or real-world agentic systems. Reducing debate depth or dynamically selecting which personas to invoke may help mitigate these costs in future work.

Overall, the goal of Multi-Agent Reflexion is to inject structured disagreement and diverse reasoning into the reflection step, reducing confirmation bias and improving the agent's ability to escape repeated failures. 

\section{Conclusions}
In this work, we introduced Multi-Agent Reflexion (MAR), a multi-agent extension of the Reflexion framework designed to address key failure modes in single-agent self-improvement. Most notably cognitive entrenchment and the degeneration-of-thought problem. Across both reasoning (HotPotQA) and program synthesis (HumanEval-Python), MAR consistently outperforms the GPT-3.5 baseline and single-agent Reflexion, demonstrating that diversified critique and structured disagreement substantially enhance error diagnosis and correction.


Looking forward, several directions appear promising: experimenting with MAD paradigms that reduce context lengths, as demonstrated by \cite{chen:25_sid} and \cite{smit:24_mad}. New discoveries that have been shown to improve LLM output diversity, such as \cite{zhang:25_verbalized_sampling} also show promise when integrated with our framework. Overall, our findings suggest that multi-agent reflective systems offer a scalable, training-free path toward more reliable and self-correcting LLM agents. 

\subsection{Societal Impact}

Our work highlights both the potential benefits and risks of developing self-improving LLM agents. Multi-agent debate offers a scalable way to elicit diverse reasoning and richer critiques without gradient-based training, suggesting a path toward more transparent and interpretable AI systems. However, our experiments also reveal notable societal concerns. Because our multi-agent Reflexion pipeline required roughly 300–400 API calls per task, around three times the cost of single-agent Reflexion, it substantially increases computational demand, energy consumption, and therefore environmental impact, while also raising economic and accessibility barriers. Moreover, self-refining agents that generate and act on their own critiques introduce safety and governance challenges, as emergent multi-agent behaviors may become harder to predict or align. By demonstrating both the promise and the heavy computational footprint of multi-agent reflexive systems, our work underscores the need for careful evaluation of efficiency, reliability, and the broader societal consequences of deploying increasingly autonomous LLM agents.


\bibliography{aaai25}

\clearpage
\onecolumn
\section{Appendix}

\subsection{A) Failure Case: Hallucinated Specification Drift in Reflexion}
\label{app:double_the_difference}

One recurring failure mode we observed in our HumanEval runs was that Reflexion sometimes \emph{hallucinated a new task specification} and confidently steered the agent away from the true objective. Listing~\ref{lst:double_the_difference_failure} shows a representative example
for the \texttt{double\_the\_difference} problem. The original prompt asks the model to return the sum of squares of positive odd integers in the list, ignoring negatives and non-integers. However, after a failed attempt, the reflection step incorrectly infers that the function is supposed to ``double the difference of the elements in the list'' and rewrites the entire
implementation accordingly. This reflection not only fails to diagnose the real bug, it
actively \emph{redefines} the task, causing the second attempt to diverge even further from
the benchmark specification. This illustrates how low-quality self-feedback in a single-agent
Reflexion loop can entrench errors instead of correcting them, and motivates our Multi-Agent
Reflexion extension, which introduces external evaluators and diverse reasoning personas to
counteract such self-reinforcing drift.

\begin{listing}[h]
\caption{Failure case: hallucinated task specification for \texttt{double\_the\_difference}.}

\label{lst:double_the_difference_failure}
\begin{lstlisting}[language=Python]

\end{lstlisting}
\end{listing}

\subsection{B) Failure Case: EM Rejects Semantically Correct Answer}
\label{app:em-womans-era}

One recurring limitation of the Exact Match (EM) metric is that it penalizes
superficial formatting differences even when the model's answer is
semantically correct. A representative example from our MAR HotPotQA runs is
the following item:

\paragraph{Question.}
\textit{``Woman's Era and Naj are what kind of magazines?''}

\paragraph{Ground-truth answer.}
\texttt{fortnightly women interest magazine}

\paragraph{MAR predictions (Trials 1--5).}
\begin{verbatim}
"women's interest magazines"
"women's interest magazines"
"women's interest magazines"
"women's interest magazines"
"women's interest magazines"
\end{verbatim}

All five trials are marked incorrect under EM, despite the fact that
``women's interest magazines'' is:

\begin{itemize}
    \item semantically identical to the core label ``women interest magazine'',
    \item factually correct for both \textit{Woman's Era} and \textit{Naj},
    \item only missing the non-essential temporal modifier ``fortnightly'',
    \item and differing only in surface form (pluralization and wording).
\end{itemize}

\noindent Importantly, inspection of the debate transcripts shows that the
reasoning is correct: MAR consistently identifies the category as women's
interest magazines and does not hallucinate an alternative type. The failure
arises entirely from EM's strict string-matching requirement rather than from
a reasoning error.

This illustrates the broader issue mentioned in the main text: these
superficial mismatches arise even when the answer matches human expectations.
Under a more forgiving metric (e.g., F1, or a semantic-match / LLM-judge
scoring scheme), this example would be counted as correct, and the apparent
gap between MAR and the original Reflexion results would likely shrink in
favor of MAR.

\subsection{C) Persona Prompts: All Prompts Used for Any Persona in Paper}

\subsection{Reasoning Personas for HotPotQA}

\paragraph{Verifier}
You are a Verifier. Your job is to check each claim carefully for factual correctness and internal logical consistency. For each assertion the Actor or another agent makes, ask yourself: “Is this backed by evidence or context? Could this be wrong?” Discard any reasoning steps that lack justification, ambiguous references, or unsupported assumptions. If you find an error or gap, call it out explicitly and explain why it might be wrong.

\paragraph{Planner}
You are a Planner. You care about the high-level structure of the reasoning. Before diving into low-level details, outline a strategy: what steps to take, what subproblems to solve, and in what order. If earlier attempts failed, propose a different overall plan (alternative breakdown). Your reflections should focus on planning, not just individual mistakes.

\paragraph{Skeptic}
You are a Skeptic. Assume earlier reasoning may have hallucinations or leaps. Critique every assumption, spec, and inference. Ask: “How do I know this is true?”, “What if the premise is wrong?”, “Is there another possible interpretation?” Your goal is to prevent overconfidence and surface plausible failure modes.

\paragraph{Logician}
You are a Strict Logician. Evaluate whether the answer exactly matches the specification or asked question. Do not accept vague matches, implied meanings, or partially correct statements. If the requirement asks for a full definition, full proof, or exact formatting, check strictly for compliance.

\paragraph{Creative}
You are a Creative Thinker. If conventional reasoning fails or stalls, propose unforeseen angles. Look for edge cases, trick questions, alternative interpretations, or unusual solutions. Your reflections should expand the search space rather than refine within the existing pattern.

\paragraph{Meta-Reflector}
You are a Meta-Reflector. After seeing multiple failed attempts, reflect not just on code or reasoning errors but on the overall process. Ask: “Why did we keep failing?”, “Are we stuck in a loop of similar mistakes?”, “Should we change the memory buffer, retry policy, or strategy type?” Suggest meta-changes: different prompting style, more memory, switching reasoning mode, or abandoning this approach.

\subsection{Programming Personas for HumanEval}

\paragraph{Senior Engineer}
You are a Senior Software Engineer. Your goal is to write clean, efficient, and correct code that passes all unit tests.

\paragraph{QA Engineer}
You are a QA Test Engineer. You focus on edge cases, input validation, and making sure the implementation truly satisfies the specification.

\paragraph{Algorithm Expert}
You are an Algorithm Specialist. Focus on correctness and time/space complexity of the underlying algorithm.

\paragraph{Code Reviewer}
You are a Strict Code Reviewer. You check for logical bugs, syntax errors, Pythonic style, and maintainability.

\subsection{D) Parameters for Extension}
\subsection{HotPotQA:}
\begin{itemize}
    \item \textbf{Number of debaters:} 4 (\textit{Skeptic}, \textit{Logician}, \textit{Creative}, \textit{Verifier})
    \item \textbf{Debate rounds per failure:} 2
    \item \textbf{Maximum trials per question:} 5 (initial attempt + up to 4 MAR-guided retries)
\end{itemize}

\subsection{HumanEval:}
\begin{itemize}
    \item \textbf{Number of debaters:} 3 (\textit{Senior Engineer}, \textit{QA Engineer}, \textit{Code Reviewer})
    \item \textbf{Debate rounds per failure:} 2
    \item \textbf{Maximum trials per problem:} 3 
\end{itemize}

\subsection{E) Poor EM reward signal from HotPot leads Reflexion away from correct response} 
In this section, we can see the reasoning trial of MAR Reflexion attempting a HotPotQA question. In the first round, it answers effectively correctly, but its answer 'Stone Brewing Co' doesn't exactly match 'Stone Brewing', causing it to be labeled incorrect and for the self-reflection process to start. On trial 5 we can see that the result of these repeated reflections is that the model keeps searching for the 'correct' response, resulting in it timing out and being ultimately incorrect. This example highlights the issue with evaluating on the HotPotQA dataset using exact string matching.\\

\textbf{Trial 1:}

\paragraph{Question:} Beer Wars covers the differences between large corporate breweries, and small breweries, such as what brewery that is headquartered in Escondido, california?

\paragraph{Thought 1}: I need to search Beer Wars and find the differences between large corporate breweries and small breweries, then find the brewery headquartered in Escondido, California.\\
Action 1: Search[Beer Wars]

\paragraph{Thought 2}: The passage mentions small breweries like Dogfish Head Brewery, Moonshot 69, Yuengling, and Stone Brewing Co. I need to find which brewery is headquartered in Escondido, California.\\
Action 2: Lookup[Escondido, California]

\paragraph{Thought 3}: I need to search Stone Brewing Co. to find out if it is headquartered in Escondido, California.\\
Action 3: Search[Stone Brewing Co.]

\paragraph{Thought 4}: Stone Brewing Co. is headquartered in Escondido, California. So the answer is Stone Brewing Co.
Action 4: Finish[Stone Brewing Co.]\\
Observation 4: Answer is INCORRECT\\
Correct answer: Stone Brewing\\

\textbf{Trial 5:}
\paragraph{Thought 1}: I need to search Beer Wars, find the differences between large corporate breweries and small breweries, and then find the brewery headquartered in Escondido, California.\\
Action 1: Search[Beer Wars]

\paragraph{Thought 2}: Beer Wars covers the differences between large corporate breweries and small breweries like Dogfish Head Brewery, Moonshot 69, Yuengling, and Stone Brewing Co. I need to find which brewery is headquartered in Escondido, California.\\
Action 2: Search[Escondido, California brewery]

\paragraph{Thought 3}: The passage does not mention the specific brewery headquartered in Escondido, California. I need to search for breweries specifically located in Escondido, California.\\
Action 3: Search[Breweries in Escondido, California]

\paragraph{Thought 4}: I need to search for breweries specifically located in Escondido, California to find the brewery mentioned in the question. Let's try searching for specific breweries in Escondido, California.\\
Action 4: Search[Escondido California breweries list]

\paragraph{Thought 5}: The search for the specific brewery headquartered in Escondido, California has not been successful so far. I should try to search for a list of breweries specifically located in Escondido, California to find the answer to the question. Let's continue searching for Escondido, California breweries list.\\
Action 5: Search[Escondido California breweries list]

\paragraph{Thought 6}: The search for the specific brewery headquartered in Escondido, California has not been successful. I should try to refine my search terms to find a more specific list of breweries in Escondido, California. Let's try searching for "Escondido California breweries list" again with a different search engine or source to find the answer to the question. Action 6: Search[Escondido California breweries list]\\
Action 6: Search[Escondido California breweries]\\

TIMEOUT: Only 6 actions allowed!\\
Correct answer: Stone Brewing




\end{document}